\documentclass[11pt,a4paper]{article}
\usepackage[hyperref]{emnlp2018}
\usepackage{times}
\usepackage{latexsym}

\usepackage{url}
\usepackage{times}
\usepackage{xcolor}
\usepackage{soul}
\usepackage[utf8]{inputenc}
\usepackage[small]{caption}
\usepackage{graphicx}
\usepackage{amsmath}
\usepackage{algorithm}
\usepackage[]{algpseudocode}

\aclfinalcopy 

\title{A Semi-Supervised Approach for Low-Resourced Text Generation}

\author{Hongyu Zang and Xiaojun Wan\\
  Institute of Computer Science and Technology, Peking University \\
  The MOE Key Laboratory of Computational Linguistics, Peking University \\
  {\tt \{zanghy, wanxiaojun\}@pku.edu.cn}
}

\date{}

\begin{document}
\maketitle
\begin{abstract}
Recently, encoder-decoder neural models have achieved great success on text generation tasks. However, one problem of this kind of models is that their performances are usually limited by the scale of well-labeled data, which are very expensive to get. The low-resource (of labeled data) problem is quite common in different task generation tasks, but unlabeled data are usually abundant. In this paper, we propose a method to make use of the unlabeled data to improve the performance of such models in the low-resourced circumstances. We use denoising auto-encoder (DAE) and language model (LM) based reinforcement learning (RL) to enhance the training of encoder and decoder with unlabeled data. Our method shows adaptability for different text generation tasks, and makes significant improvements over basic text generation models.
\end{abstract}

\section{Introduction}
Text generation is a central task of natural language processing (NLP). Models are designed to generate target natural language texts from various kinds of source data, for example, data records, texts, voices and pictures.


With the fast development of neural network techniques, text generation has achieved a historical leap in the past few years. For example, there are several breakthroughs happening in neural machine translation tasks in recent years \cite{sutskever2014sequence,wu2016google}, generative document summarization models achieve competitive performance \cite{TanWX17}, and cross-modal tasks like image captioning also get impressive progress with neural models \cite{VinyalsTBE15}. These works are commonly based on the popular encoder-decoder framework. Such models learn from labeled source-target data pairs (i.e., parallel data), and end up with an encoder (which maps the source-side data into hidden features), and a decoder (which maps the hidden features into target-side data). Models with adequate parameters are powerful enough to learn the mappings from the training data and get good generalization capability. And usually with more data, the models become more powerful, as they can explore more of the data and learn the mapping more accurately from training data.

Here comes the problem that labeled data are not easy to get. Although a lot of works \cite{DBLP:conf/sigmod/2015,DBLP:conf/acl/GardentSNP17} have been done to alleviate the problem by using machine to automatically detect pairwise data or to help human to do that, these methods are mainly task-specified, and limited to certain domains. Automatically getting a lot of labeled data is still impossible for most circumstances. The lack of data causes neural models not performing well in many tasks. However, there are tons of unlabeled data left with no use, which is totally a waste. It is not easy for human-beings to learn speaking from zero without any labeled data. But with some labeled ones, we can make use of the unlabeled data. So do the machines. And inspired by this, we develop a semi-supervised approach to make use of both labled and unlabeled data, letting the labeled ones guide the training procedure and the unlabeled ones make enhancement.

The main purpose of this work is to explore how to use the unlabeled data to enhance the training process of encoder-decoder text generation models. Our principle is to keep it simple, reasonable, and effective.

Our method is like a scaffold. Apart from the basic model serving as the core, we use the DAE (Denoising Auto-Encoder) to help train the decoder part with unlabeled target-side data. And we use LM (Language Model) based RL (Reinforcement Learning) to put emphasis on helping explore source-side data with unlabeled data. We share the encoder and decoder's parameters among different training strategies. And finally, we can get a same encoder-decoder network just as the basic model, but with better parameters.

We designed models and conducted experiments on two different text generation tasks. We applied extensive evaluations and comparisons in our experiments. The results demonstrate that our method does help improve the performance of basic models under low-resourced circumstances in different tasks.

Our main contributions can be concluded as follows:

\begin{itemize}
	\item We address the problem of how to reinforce the encoder and decoder with unlabeled source-side data and target-side data by using DAE and RL.
	\item We propose a reinforcement reward function for general text generation tasks.
	\item The method we propose offers a new way for semi-supervised learning with encoder-decoder text generation models. Experiments demonstrate that it has adaptability for different text generation tasks. And  it also shows the potential to be applied to more general cases.
\end{itemize}

\section{Preliminaries}
\subsection{Basic Model}
In this paper, we mainly experiment on text-to-text generation models. So we regard the popular sequence-to-sequence model with attention \cite{bahdanau2014neural} as our basic model.
\subsubsection{Recurrent Neural Network (RNN)}
RNN is designed to generate hidden states $H={h_1,h_2,h_3,...h_{|X|}}$ according to a sequence of inputs $X={x_1,x_2,x_3,...,x_{|X|}}$. Usually, traditional RNN unit will suffer from long-term information loss heavily, which we call gradient vanishing problem. Cell structures like LSTM \cite{hochreiter1997long} and GRU \cite{chung2014empirical} are designed to address this problem. In our work, we use LSTM as the basic RNN unit.


%
%
%

\subsubsection{Sequence-to-Sequence Model}
Sequence-to-sequence model is a popular encoder-decoder model for tasks requiring transformation from inputs $X$ to outputs $Y={y_1,y_2,y_3,...,y_{|Y|}}$ \cite{sutskever2014sequence}. It uses RNN as encoder and decoder. The encoder computes a series of hidden states $H$ as outputs, which represent high-dimensional features of the inputs $X$. And then the decoder takes $H$ and additional inputs $X'={x'_1,x'_2,x'_3,...,x'_{|Y|}}$ to compute the output hidden states $H'={h'_1,h'_2,h'_3,...h'_{|Y|}}$. Some techniques like softmax are used to transform the decoder outputs into a sequence of probability distributions $P={p_1,p_2,p_3,...,p_{|Y|}}$ to sample an output sequence $Y'={y'_1,y'_2,y'_3,...,y'_{|Y|}}$. This procedure is described by Equations (1)(2)(3).

And for the training process, all the parameters are usually updated by minimizing the cross entropy loss according to $P$ and $Y$ by Equation (4).

\begin{equation}
H = {RNN}_{encoder}(X)
\end{equation}

\begin{equation}
p_t = softmax({RNN}_{decoder}(x'_t, H))
\end{equation}

\begin{equation}
y'_t = argmax_w(p_{t,w})
\end{equation}

\begin{equation}
Loss_{CE} = -\frac{1}{|Y|}\displaystyle\sum_{t=1}^{|Y|}log(p_{t,y_t})
\end{equation}

\subsubsection{Attention Mechanism}
Attention mechanism for sequence-to-sequence model is proposed by \cite{bahdanau2014neural}. It allows the decoder to find information of alignments (contexts) before decoding. The context vector $c$ at a specific time $t$ is calculated as Equations (5)(6)(7). The attention mechanism we use in this paper is Luong attention \cite{DBLP:conf/emnlp/LuongPM15}.

\begin{equation}
s_{t,j} = score(h'_{t-1}, h_{j-1})
\end{equation}

\begin{equation}
a_{t,j} = \frac{\bf{e}^{s_{t,j}}}{\sum_{k=1}^{|X|}\bf{e}^{s_{t,k}}}
\end{equation}

\begin{equation}
c_t = \displaystyle\sum_{j=1}^{|X|}a_{t,j}h_j
\end{equation}

\begin{figure*}[h!]
    \centering
    \includegraphics[width=5in]{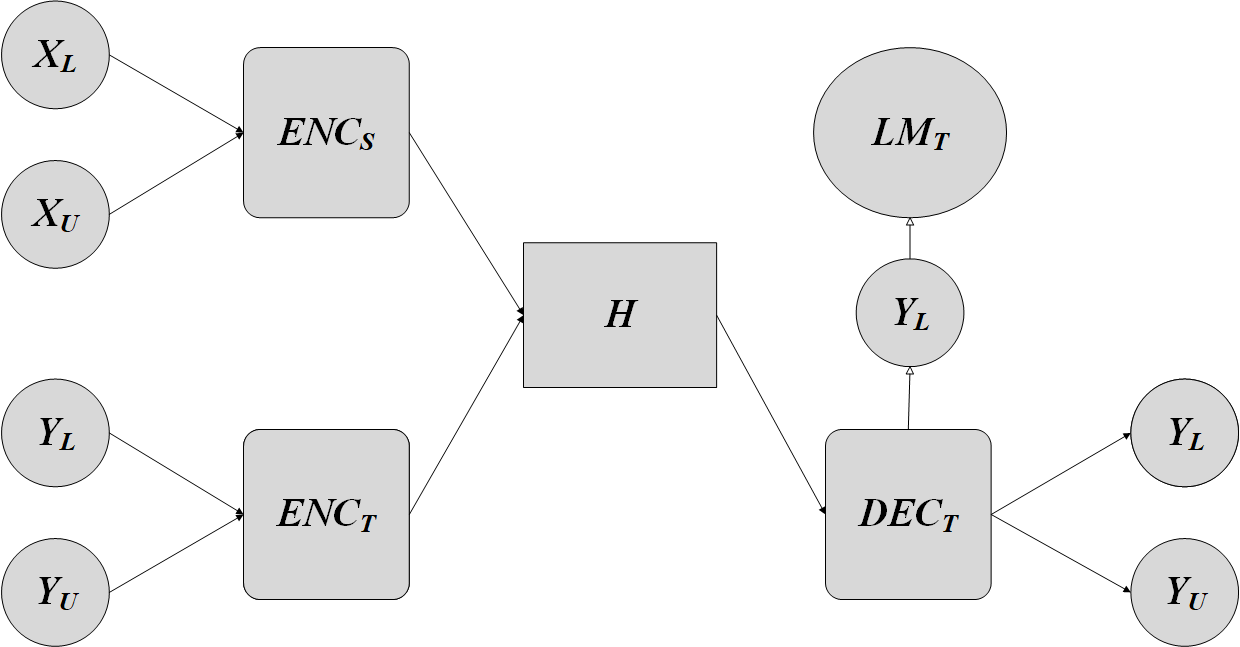}
    \caption{The model structure of our proposed method. The hollow arrows suggest that reinforcement learning may also be applied to any route.}
    \label{figures}
\end{figure*}

\subsection{Policy Gradient}
Policy gradient is a simple but widely-used algorithm for RL. In text generation, we regard the whole network as a stochastic policy, the prediction as actions and value received from some score function as reward.

Text generation tasks usually have a large action space (dictionary size). Under this circumstance, the REINFORCE algorithm \cite{williams1992simple} is used to simplify the optimization process. The algorithm regards the reward as a single-sample estimate and minimizes the objective function as Equation (8), where $r$ is the reward function.

\begin{equation}
Loss_{RL} = -\frac{1}{|Y|}\displaystyle\sum_{t=1}^{|Y|}r(y'_t| Y')log(p_{t,y'_t})
\end{equation}

\section{Unlabeled Data Reinforced Generation}
The basic model consists of two important components, the encoder and the decoder. Thus, the goal can be detailed into two parts. How to reinforce the encoder? And how to reinforce the decoder? On the other hand, there are yet two another important questions for model design. How to use the unlabeled source-side data? And how to use the unlabeled target-side data? To explore answers to these questions, we come up with a method using DAE, RL and shared parameters.

Figure 1 shows how we apply our method to the basic encoder-decoder text generation model. There are three main modules, source-side data's encoder ($ENC_{S}$), target-side data's encoder ($ENC_{T}$), and target-side data's decoder ($DEC_{T}$). These modules can be various neural networks, while they are all RNNs in our experiments. And the source-side data $X$ and target-side data $Y$ are subscripted by $L$ and $U$, to specify whether they are labeled (parallel) or unlabeled. There is a unique module called $LM_{T}$, which plays the role of LM scorer (reward function) for target-side texts. In a word, there are 3 important routes sharing parameters. These routes are designed with different motivations and they are jointly trained to make a better encoder-decoder generation model.

Apparently, if we have some labeled data (i.e., source-target parallel data), we should make the most of them. As we do not focus on how to use labeled data more effectively, we just use them to train a basic generation model. This is corresponding to the route of $X_L\rightarrow ENC_{S}\rightarrow H\rightarrow DEC_{T}\rightarrow Y_L$, which is denoted as $ROUTE \#1$.

As for the unlabeled target-side data, fortunately, DAE is a popular unsupervised model to train a decoder, which intends to reproduce the correct inputs from noisy inputs. We use the unlabeled target-side data to train a DAE, and it shares the decoder with the basic model in $ROUTE \#1$, expecting to train a decoder with better capacity. This is $ROUTE \#2$, propagating forwardly as $\{Y_L, Y_U\}\rightarrow ENC_{T}\rightarrow H\rightarrow DEC_{T}\rightarrow \{Y_L, Y_U\}$.

Finally, $ROUTE \#3$ is designed as $\{X_L, X_U\} $ $\rightarrow ENC_{S}\rightarrow H\rightarrow DEC_{T}\rightarrow Y' \rightarrow LM_{T}$. With $ROUTE\#2$ having already enhanced the decoder, we think the weakness in encoder could lie in that the data sparsity of labeled data hinders the encoder to explore the source-side data. For the unlabeled source-side data, we encode and decode them with the basic generation model. But there are no labeled results for the model's output to compute the cross entropy loss and update parameters. Therefore, we apply the reinforcement learning algorithm with a reward function drawn by LM in the target language to give a reward for optimization. The impact of the RL reward on the decoder may be weak, but it provides a way to learn from unlabeled source-side data, which will support a better exploration of source-side data. While we have to use RL to train $ROUTE\#3$, the RL loss can also be applied to $ROUTE\#1$ and $ROUTE\#2$ to encourage the model to produce outputs that get higher scores from the reward function.

The whole training procedure is shown in Algorithm 1. Finally, we can obtain an enhanced encoder and an enhanced decoder for text generation. More details and discussions are presented in the following subsections.

\begin{algorithm}[t!]
\caption{Training Procedure}
\label{pseudocode 1}

\begin{algorithmic}[1]
    \Require \textit{Data}:$X_L$, $X_U$, $Y_L$, $Y_U$; \textit{Denoise Function}:$D$; \textit{Language Model}:$LM_T$; \textit{Boolean}:$ALL\_USE\_RL$.
    \Ensure \textit{Encoder-Decoder Model}: $ENC_{S}$, $DEC_{T}$.

    \Procedure{GET\_RL\_LOSS}{$Y'$, $n$}
        \If {$ALL\_USE\_RL$ \textbf{or} $n=3$}
            \State \textbf{Return} $REINFORCE(LM_{T}, Y')$
        \Else
            \State \textbf{Return} $0$
        \EndIf
    \EndProcedure
    \Procedure{TRAIN\_ROUTE\#1}{$X$, $Y$}
        \State $P, Y' \gets DEC_{T}(ENC_{S}(X), Y)$
        \State $Loss_{CE} \gets CrossEntropy(P,Y)$
        \State $Loss_{RL} \gets GET\_RL\_LOSS(Y', 1)$
        \State Update $\{ENC_{S}, DEC_{T}\}$ with $Loss_{CE}+Loss_{RL}$
    \EndProcedure
    \Procedure{TRAIN\_ROUTE\#2}{$Y$}
        \State $P, Y' \gets DEC_{T}(ENC_{T}(D(Y)), Y)$
        \State $Loss_{CE} \gets CrossEntropy(P,Y)$
        \State $Loss_{RL} \gets GET\_RL\_LOSS(Y', 2)$
        \State Update$\{ENC_{T}, DEC_{T}\}$ with $Loss_{CE}+$ $Loss_{RL}$
    \EndProcedure
    \Procedure{TRAIN\_ROUTE\#3}{$X$}
        \State $Y' \gets DEC_{T}(ENC_{S}(X), Y')$
        \State $Loss_{RL} \gets GET\_RL\_LOSS(Y', 3)$
        \State Update $\{ENC_{S}, DEC_{T}\}$ with $Loss_{RL}$
    \EndProcedure
    \State Randomly initialize$\{ENC_{S},ENC_{T},DEC_{T}\}$
    \While {Not Converged}
        \State $n \gets \textbf{RANDOM}({1,2,3})$
        \If {n=1}
            \State $TRAIN\_ROUTE\#1(X_L, Y_L)$
        \ElsIf {n=2}
            \State $TRAIN\_ROUTE\#2(Y_L \cup Y_U)$
        \Else
            \State $TRAIN\_ROUTE\#3(X_L \cup X_U)$
        \EndIf
    \EndWhile

\end{algorithmic}
\end{algorithm}

\subsection{Denoising Auto-Encoder}
Basic auto-encoder (AE) targets at restoring the inputs. Such model trained by an attentive sequence-to-sequence model can be difficult to transfer to other tasks, because it is much easier for the model to learn to copy than to properly encode and decode. As many previous works \cite{DBLP:journals/corr/abs-1710-11041,DBLP:journals/corr/abs-1711-00043,DBLP:conf/acl/SennrichHB16} suggest, some noise can be added to the original inputs. The improved model is denoising auto-encoder.

In our work, we apply a simple way to add noise to the inputs. As our inputs are texts, for every word in an input sentence, there is 10\% chance that we delete it, 10\% chance that we duplicate it, and 10\% chance that we swap it with the next word, or it remains unchanged. Note that We only add noise to the inputs for $ROUTE \#2$.

\subsection{Reinforcement with Language Model}
LM is a simple but effective way to compute the prior probability of a certain sequence in a specific language. As LM is an unsupervised statistical model, it is a good option to make use of the unlabeled data. However, LM is not an ideal model for text generation, as it cannot generate texts according to specific inputs. We develop a reward function in Equation (9) based on the LM probabilities $PL$. This reward function computes for every output word, using its possible 3-grams' possibilities. It reflects how likely the model's outputs play a positive role in composing sentences in the target language.
%

\begin{equation}
r(y'_t| Y') = Mean\left\{ \begin{array}{l} ln(PL(y'_{t-2}, y'_{t-1}, y'_t)) \\ ln(PL(y'_{t-1}, y'_t, y'_{t+1})) \\  ln(PL(y'_t, y'_{t+1}, y'_{t+2})) \end{array} \right\}
\end{equation}

In general case, the loss function we need to optimize is Equation (10), where $\alpha$ is a hyper-parameter to control the ratio of different losses. And the RL part will be optimized according to the REINFORCE algorithm.

\begin{equation}
Loss = Loss_{CE} + \alpha*Loss_{RL}
\end{equation}

\subsection{Discussions on Managing Encoder and Decoder}
It seems that we have more options for our design. For instance, we can share the encoder between source-side and target-side data, and we can also add more routes like $\{X_L, X_U\}\rightarrow ENC_{S}\rightarrow H\rightarrow DEC_{S}\rightarrow \{X_L, X_U\}$. But according to our experiments, the results of such designs are negative. And we would like to discuss about that.

For sharing encoder, the source-side data and target-side data are usually considered as different linguistic forms, or even different modalities for more general situations. There are so many differences between them. Although cross-lingual or cross-modal embeddings can be used, they can not address the structure differences (grammar, order) of the data.

For adding more routes, it means two unique decoders will have to share a common hidden space. But for different decoders, the differences between the required hidden spaces can be totally different, which cannot be ignored. Some technics like VAE\cite{DBLP:journals/corr/KingmaW13} and  GAN\cite{goodfellow2014generative} can be applied to control the hidden spaces to follow some distributions, but they will not solve the problem perfectly.

To make things simple, in this work, there will not be any shared encoder or decoder across the source-side and target-side data. But it does not mean sharing will not work with some advanced techniques \cite{DBLP:journals/corr/abs-1710-11041,DBLP:journals/corr/abs-1711-00043}.

\section{Experiments}
\subsection{Tasks and Datasets}
\subsubsection{Data-to-Text Generation on WebNLG}
The WebNLG dataset is a corpus designed for data-to-text generation \cite{DBLP:conf/acl/GardentSNP17}. And there was an open challenge organized at the INLG conference based on this corpus\footnote{http://webnlg.loria.fr/pages/challenge.html}. The released training set contains 18102 data pairs. And the official development set has 2268 data pairs. The source-side data of this corpus is a set of triples like \textit{John birthDate 1942\_08\_26}, and the target is some texts describing the semantic of the triple set like \textit{John was born on 1942-08-26}. There are at most 7 relevant triples for each data pair, which means the input can be really complex. As we can transform the triples to plain texts, the tasks can be regard as a text-to-text generation task. The average length of the target text is 24 words.

The organizer has published the training set and development set for the WebNLG challenge. But the test set is not public. Therefore, in our experiments, we use the published development set as our test set. And we sampled 1000 data pairs from the training set as our development set.
\subsubsection{Sentence Simplification on Newsela}
Newsela is a corpus targeting at the sentence simplification task. It is a multi-version news corpus collected by editors from the newsela company\footnote{https://newsela.com} and collated by \cite{DBLP:journals/tacl/XuCN15}. Xu \shortcite{DBLP:journals/tacl/XuCN15} argued that there are several shortcomings in the Simple Wikipedia dataset limiting simplification researches, while Newsela is of higher quality. We quite agree with the points, and we choose to use Newsela rather than more resourced Wikipedia.

In previous work, Zhang and Lapata \shortcite{DBLP:conf/emnlp/ZhangL17} used Newsela to develop an RL approach optimizing the simplification model. We use their preprocessed data in our experiments. In this dataset, we have 94208 training pairs, whose input texts are more complex than output texts. There are another 1129 complex-simple pairs for development and 1077 pairs for test.

\begin{table*}[t!]
\begin{center}
\begin{tabular}{|c|c|c|c|c|c|}
\hline\bf {Models}&\bf {BLEU}&\bf {METEOR}&\bf {TER}&\bf {PPL}&\bf {HUMAN}\\\hline
\bf{R\#1}&42.82&21.28&63.04&55.15&3.02\\\hline
\bf{R\#1+LM}&42.23&\underline{22.80}&61.58&52.98&\underline{3.34}\\
\bf{R\#1,2+LM}&\underline{43.59}&\bf\underline{24.10}&\bf58.08&\bf\underline{30.57}&\underline{3.51}\\
\bf{R\#1,2,3+LM}&\bf\underline{48.04}&\underline{23.03}&59.11&\underline{31.18}&\bf\underline{3.69}\\
\hline
\end{tabular}
\end{center}
\caption{\label{tables}Model comparison results on WebNLG. \textbf{R} is the abbreviation of ROUTE. The numbers after \textbf{R} specify which ROUTEs are applied. \textbf{+LM} means the RL reward based on LM is used to reinforce training for all routes. Results significantly better ($p<0.01$ in a two-tail T-test) than basic model ($R\#1$) get underlined.}
\end{table*}

\begin{table*}[t!]
\begin{center}
\begin{tabular}{|c|c|c|c|c|c|}
\hline\bf {Models}&\bf {BLEU}&\bf {SARI}&\bf {FKGL}&\bf {PPL}&\bf {HUMAN}\\\hline
\bf{R\#1}&14.15&33.92&2.13&62.18&2.06\\\hline
\bf{R\#1+LM}&14.36&34.14&2.15&60.34&\underline{2.27}\\
\bf{R\#1,2+LM}&16.23&\underline{34.40}&\bf\underline{1.84}&54.60&\underline{2.32}\\
\bf{R\#1,2,3+LM}&\bf\underline{17.32}&\bf\underline{35.32}&2.25&\bf\underline{52.46}&\bf\underline{2.66}\\
\hline
\end{tabular}
\end{center}
\caption{\label{tables}Model comparison results on Newsela.}
\end{table*}

\subsection{Training Details}
We use Tensorflow 1.2\footnote{https://github.com/tensorflow/tensorflow/tree/r1.2} for implementation. As both the tasks can be regarded as text-to-text generation, we use LSTM cells for bidirectional RNN encoder and RNN decoder with Loung attention. The dimension sizes of embedding vectors and hidden vectors are all set to 500. And for the output of RNN cells, we use a dropout rate of 0.3. Models are joint trained as mentioned in Section 3. We use the gradient descent optimizer to optimize models with a learning rate of 1.0. As for building language models, we use IRSTLM\footnote{https://github.com/irstlm-team/irstlm}, setting $n=3$ (n-gram size), and $k=1$ (number of splits).

Specially, for WebNLG experiments, the encoder is one-layer, while the decoder is two-layer. The ratio of the RL loss is $\alpha=0.2$. As for Newsela experiments, both the encoder and decoder are two-layers, with $\alpha=1.0$.

\subsection{Evaluation Metrics}
For WebNLG, we use BLEU\footnote{https://github.com/moses-smt/mosesdecoder/blob/ master/scripts/generic/multi-bleu.perl}\cite{DBLP:conf/acl/PapineniRWZ02}, METEOR\footnote{http://www.cs.cmu.edu/~alavie/METEOR/}\cite{denkowski:lavie:meteor-wmt:2014}, TER\footnote{http://www.cs.umd.edu/~snover/tercom/}\cite{snover2006study}, and perplexity (PPL) as automatic evaluation metrics. BLEU, METEOR and TER are recommended by WebNLG Challenge organizers. PPL is a transformation of cross entropy loss on development set and test set. Better outputs are expected to get higher BLEU and METEOR scores, and lower TER and PPL. Apart from automatic evaluation, we also apply human evaluation to evaluate the usability of the outputs. In human evaluation, we sample 50 data inputs from the test set and obtain the output texts of different models. These data pairs are presented to 5 workers via Amazon Mechanical Turk (AMT\footnote{https://www.mturk.com/}). They are asked to rate the output texts according to the input triple set using a 5-pt Likert scale\footnote{https://en.wikipedia.org/wiki/Likert\_scale}. The final scores we present is the average of all the human evaluation results.

For Newsela, we use BLEU, SARI\footnote{https://github.com/cocoxu/simplification/blob/master/ SARI.py}\cite{DBLP:journals/tacl/XuNPCC16}, FKGL\footnote{https://github.com/mmautner/readability/blob/master/ readability.py}, and PPL. SARI is a specific metric designed for simplification evaluation. FKGL is used to evaluate the level of simpleness of texts. Better outputs are expected to get higher BLEU and SARI, and lower FKGL and PPL. And we also perform human evaluation with the same settings as WebNLG.

\subsection{Model Comparison}
We design comparison experiments to compare the performance of different models with the same data settings.

For WebNLG, we randomly sample 500 pairs of labeled data, and 14000 unlabeled source and target data items each. And For Newsela, we sample 20000 pairs of labeled data and 70000 unlabeled data items for input and output separately. The results are shown in Table 1\footnote{T-test is not performed on TER, because the tool does not offer a sentence level interface.} and Table 2.

As can be seen in the results, our method significantly improves the basic models in both automatic evaluations and human evaluations. And the comparisons of results between neighbors in the tables show the improvement can be made by every part of our method. According to the comparisons between \textit{R\#1} and \textit{R\#1+LM} in both tables, the RL reward's help is limited when applied to model training with labeled data. But it shows good impact on the training process with unlabeled source data (see \textit{R\#1,2+LM} and \textit{R\#1,2,3+LM}), which supports our belief of helping explore the source-side data. And comparing \textit{R\#1+LM} and \textit{R\#1,2+LM}, we can see that the DAE part seems to help a lot with the performance, and we believe our way to use unlabeled target data to reinforce the training of decoder is a success. To get a clearer view, we present some samples from the test set in Figure 2. Generally speaking, we consider model \textit{R\#1,2,3+LM} as the best one and BLEU as the most consistent metric with human evaluation.

\begin{figure}[t!]
    \centering
    \includegraphics[width=3in]{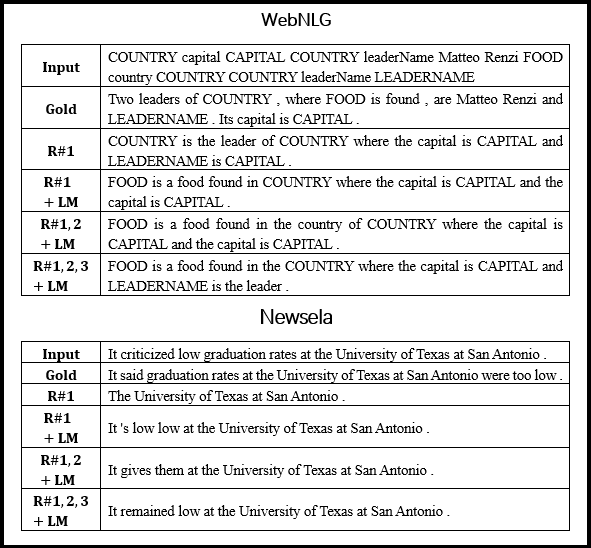}
    \caption{Comparison of different model outputs.}
    \label{figures}
\end{figure}

\subsection{Performance with Different Data Scales}
Now we compare models with different data settings to explore how the scale of labeled data and unlabeled data impacts on the improvement of performance using our method \textit{R\#1,2,3+LM}.
\subsubsection{Different Scales of Labeled Data}
For WebNLG, we use fixed 14000 unlabeled data items in source side and target side, and randomly sample labeled data with different data scales (number of data pairs) within \{200, 500, 1000, 2000\}. And for Newsela, we use fixed 70000 unlabeled data items, and the scale of labeled data is within \{1000, 5000, 10000, 20000\}.

From Figure 3, we can easily draw a conclusion that the more labeled data we have, the better performance for all models we get. Besides, our method always improves the basic model (averagely 2.19 BLEU score).

\begin{figure}[t!]
    \centering
    \includegraphics[width=3in]{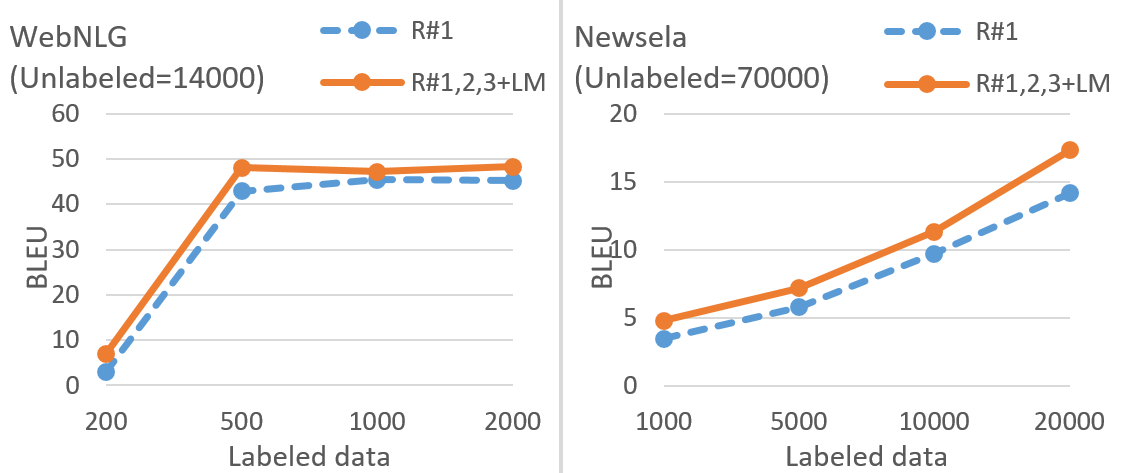}
    \caption{The comparisons of BLEU performance between basic models and our method with different scales of labeled data.}
    \label{figures}
\end{figure}
\begin{figure}[t!]
    \centering
    \includegraphics[width=3in]{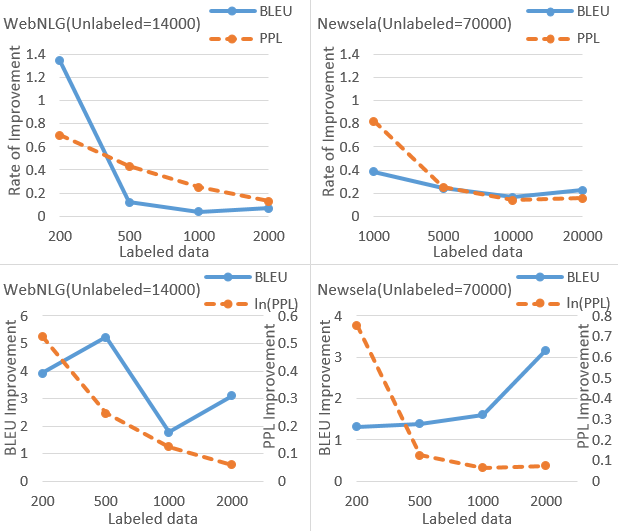}
    \caption{The analysis of BLEU and PPL improvements over basic models with different scales of labeled data.}
    \label{figures}
\end{figure}

However, as shown in Figure 4, there is also a general tendency that with more labeled data, our method helps less for the basic model. But it is reasonable, because if we have enough labeled data to train a model, unlabeled data may be useless. Another thing we should note is that when there are very few labeled data, our method is not able to totally change the situation of models giving really bad results, but still improves the basic models a lot. It does not violate the motivation of our design, as we propose the method to reinforce the basic model, and we do not expect it to learn from zero-shot.

\subsubsection{Different Scales of Unlabeled Data}
Labeled data used in these experiments is the same as that used for model comparisons. The scale of unlabeled data for WebNLG is chosen from \{1000, 5000, 10000, 14000\}, while for Newsela, the scale is chosen from \{5000, 10000, 30000, 70000\}.

\begin{figure}[t!]
    \centering
    \includegraphics[width=3in]{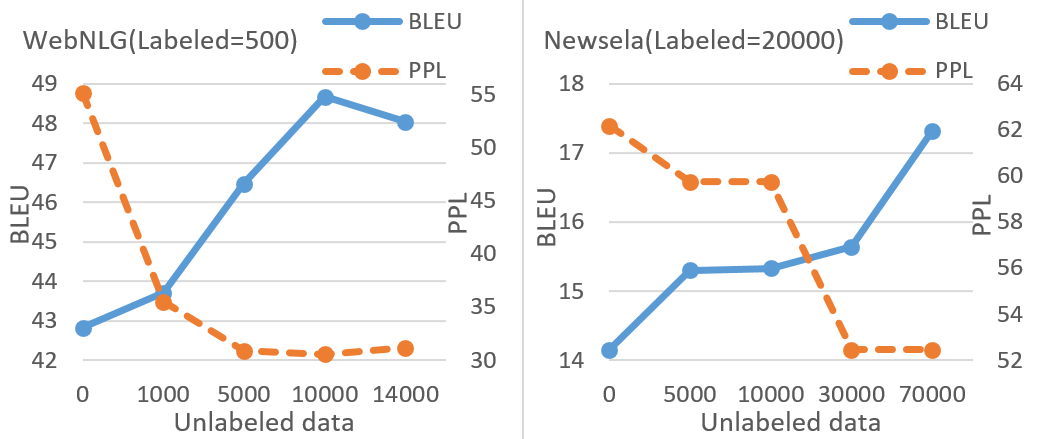}
    \caption{Comparisons among circumstances with different scales of unlabeled data.}
    \label{figures}
\end{figure}

As shown in Figure 5, even small amount of unlabeled data can play a positive role in improving the performance. In a way, more unlabeled data may be of more help. However, in the experiment of WebNLG, the model with 10000 unlabeled data is even a little bit better than the one with 14000 data. The reason may be the uncertainty of experiments, the limit of our RL reward function, and so on. But we should also come up with a guess that a certain number of unlabeled data may be enough for a task, where more unlabeled data could bring more noise and be not beneficial.

\section{Related Works}
Encoder-decoder models are very popular for various generation tasks. Despite many successes following the framework, we see the need to improve such models with unlabeled data when the data source is not so ideal, which makes the initial motivation of our work.

However, we are not the only researchers seeing the need. There are a lot of previous works working on semi-supervised learning. For example, Kingma \shortcite{DBLP:conf/nips/KingmaMRW14} propose two deep generative models for general semi-supervised learning and apply it to image generation and classification, Aminiand Gallinari \shortcite{DBLP:conf/sigir/AminiG02} try to improve classification models for text summarization with unlabeled data, and Sennrich \shortcite{DBLP:conf/acl/SennrichHB16} use two simple techniques to obtain substantial improvements on NMT models. As far as we know, most related works of semi-supervised learning in the NLP area are mainly focus on classification and categorization problems, and our proposed method can provide a new way to supplement semi-supervised solutions for text generation.

Two most relevant works \cite{DBLP:journals/corr/abs-1710-11041,DBLP:journals/corr/abs-1711-00043} are published recently. They target at using monolingual corpus to train NMT models. Apart from some techniques like denoising and backtranslation, Artetxe \shortcite{DBLP:journals/corr/abs-1710-11041} use a shared encoder, which refers to fixed pretrained cross-lingual word embeddings. Their idea of using cross-lingual embeddings to train a shared encoder is interesting and effective for training with unlabeled data. But their shared encoder seems to make the performance of training with labeled data lower than average, which we do not consider as a good option for semi-supervised training. The unique technique  Lample \shortcite{DBLP:journals/corr/abs-1711-00043} use is adversarial training. They try to enforce the distributions of different encoders' outputs to be the same by applying adversarial losses to the hidden vectors. However, they do not show how well their method will work for semi-supervised learning. Our work borrows some ideas from their works, like DAE. But our work neither targets at the specific NMT tasks, nor tries to use only unlabeled data for training. We intend to explore a new approach to make use of unlabeled data when the labled ones are low-resourced. All the techniques and tricks in other related works can also be applied to our approach to make a better model for a specific task.

\section{Conclusions and Future Work}
In this paper, we proposed a semi-supervised approach to address the training process encoder-decoder text generation models under low-resourced circumstances. We developed a series of comparison experiments on two different text generation tasks. The experiment results demonstrate that our method almost always helps improve the basic models significantly.

Apart from efficacy, the method we proposed is also easy to extend. As the final model is still an encoder-decoder model, many other techniques and tricks in related works can be easily applied to get better results. What's more, the encoder we have do not necessarily need to encode texts, some other tasks, like image caption and speech recognition, also seem to work well with our method. And this can be an interesting direction for the future extension of our work. We will also try to develop better reward functions to make more significant improvements with our method.

\section*{Acknowledgments}
This work was supported by Key Laboratory of Science, Technology and Standard in Press Industry (Key Laboratory of Intelligent Press Media Technology).
We thank Xingxing Zhang, Mirella Lapata, Luke Orland, Wei Xu, and the Newsela Company for their help to process the datasets. Xiaojun Wan is the corresponding author.



\bibliography{emnlp2018}
\bibliographystyle{acl_natbib_nourl}

\end{document}